\documentclass[10pt,twocolumn,letterpaper]{article}

\usepackage{cvpr}              
\definecolor{cvprblue}{rgb}{0.21,0.49,0.74}
\usepackage{hyperref}
\hypersetup{breaklinks,colorlinks,allcolors=cvprblue}
\usepackage{multirow}
\usepackage{adjustbox}
\usepackage{xspace}
\usepackage{subcaption}

\def\paperID{20286} 
\def\confName{CVPR}
\def\confYear{2026}

\newcommand{\ours}{{TACS}}
\newcommand{\oursfull}{Task-Aligned Context Selection\xspace}

\newcommand{\nocontxt}{{No-Context}}
\newcommand{\randctx}{{Random Context Retrieval}}
\newcommand{\frozenretr}{{Frozen DINO Similarity Retrieval }}
\newcommand{\featavg}{{Feature-Averaged Context}}
\newcommand{\nocontxtsm}{{No-Context}}
\newcommand{\randctxsm}{{Random Context Ret.}}
\newcommand{\frozenretrsm}{{Frozen DINO Similarity Ret.}}
\newcommand{\featavgsm}{{Feature-Averaged Context}}

\title{Learning What Helps: Task-Aligned Context Selection for Vision Tasks}

\author{Jingyu Guo \textsuperscript{
$1,2$}
\hspace{-1.5mm}
\thanks{Corresponding author: Jingyu Guo \textless{}jingyug@kth.se\textgreater{}} 
\qquad
Emir Konuk \textsuperscript{$1,2,3$}
\qquad
Fredrik Strand \textsuperscript{$3$} 
\qquad
Christos Matsoukas \textsuperscript{$4$} \\
\qquad
Kevin Smith \textsuperscript{$1,2$}
\\\\
\textsuperscript{$1$} KTH Royal Institute of Technology, Stockholm, Sweden 
\textsuperscript{$2$} Science for Life Laboratory, \\ Stockholm, Sweden 
\textsuperscript{$3$} Karolinska Institutet, Stockholm, Sweden 
\textsuperscript{$4$} Pathology, Clinical \\ Pharmacology and Safety Sciences, R\&D AstraZeneca, Gothenburg, Sweden
}

\definecolor{jimcolor}{RGB}{83, 158, 198}
\definecolor{blue_munsell}{rgb}{0.36, 0.54, 0.66}
\definecolor{blue-violet}{rgb}{0.54, 0.17, 0.89}
\definecolor{byzantine}{rgb}{0.74, 0.2, 0.64}
\definecolor{caputmortuum}{rgb}{0.35, 0.15, 0.13}
\definecolor{alizarin}{rgb}{0.82, 0.1, 0.26}
\definecolor{light_grey}{rgb}{0.6, 0.6, 0.6}

\newif\ifshowcomments

\showcommentsfalse

\begin{document}
\maketitle
\begin{abstract}

Humans often resolve visual uncertainty by comparing an image with relevant examples, but ViTs lack the ability to identify which examples would improve their predictions. 
We present \oursfull (\ours), a framework that learns to select paired examples which truly improve task performance rather than those that merely appear similar. 
\ours~jointly trains a selector network with the task model through a hybrid optimization scheme combining gradient-based supervision and reinforcement learning, making retrieval part of the learning objective.
By aligning selection with task rewards, \ours~enables discriminative models to discover which contextual examples genuinely help.
Across 18 datasets covering fine-grained recognition, medical image classification, and medical image segmentation, \ours~consistently outperforms similarity-based retrieval, particularly in challenging or data-limited settings. 



\end{abstract}    
\section{Introduction}
\label{sec:introduction}

\begin{figure}[t]
    \centering
    \includegraphics[width=0.9\linewidth]{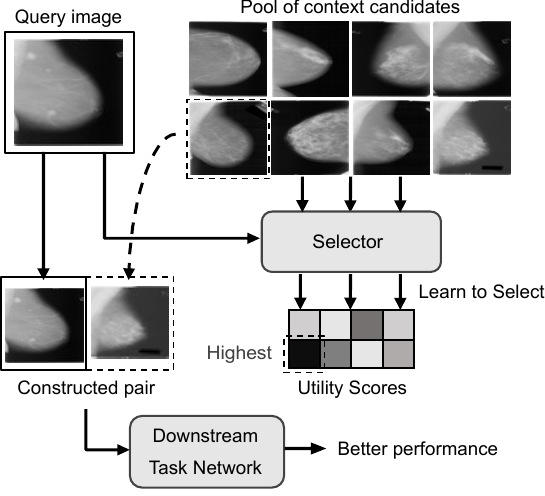}
    \caption{Overview of {\oursfull{} (\ours)}.
Given a query image, \ours{} learns to select the most informative example from a candidate pool to form a task-aligned input pair.
A \textit{selector network} predicts which candidate provides the most useful context for the \textit{downstream task network}, (\eg classifier or segmentor) which operates on the selected pair.
By learning to retrieve helpful rather than merely similar examples, \ours{} improves decision accuracy and robustness, particularly in ambiguous cases. 
}
    \label{fig:introduction}
\end{figure}

Large language models (LLMs) have evolved into multi-modal generalists that excel across a broad range of tasks: they can reason across modalities, concepts, and domains. 
They can even improve their answers by grounding their predictions through the retrieval of relevant external information~\cite{karpukhin2020dense, lewis2020retrieval, zheng2025retrieval}. 
This ability has become a defining feature of generalist intelligence, with recent works now proposing methods to {learn what to retrieve}~\cite{asai2023selfrag, gao2024smartrag}. 
Specialist models, however, currently lack mechanisms for incorporating retrieved information into their decisions apart from a few multi-view or multi-instance methods that rely on pre-defined or manually paired data~\cite{shuai2022adaptive, alzahrani2024selective, hou2024learning}.
Nevertheless, architectures such as Vision Transformers (ViTs) continue to dominate discriminative vision tasks like classification and segmentation \cite{simeoni2025dinov3, bolya2025PerceptionEncoder}, especially in specialized domains such as medical analysis \cite{nath2025vila}.
What remains unexplored is how a specialist model could {learn to retrieve} the visual context that would help it perform better. 
This highlights a critical gap: we have taught our generalists to find the information they need, but not our specialists. 
Bridging this gap raises a fundamental question: \textit{can a specialized vision model learn which contextual examples most improve its own performance?}

Human specialists like radiologists use contextual information to improve visual judgments. 
For example, when distinguishing between benign and malignant findings, a radiologist may look not only at similar prior cases but also at different ones to clarify diagnostic boundaries. 
Likewise, a visual recognition system may benefit more from pairing an input image with a complementary example, \eg, one that reveals discriminative contrasts, than from retrieving the nearest neighbor. 
This observation suggests that learning to select \emph{helpful} examples, rather than merely \emph{similar} ones, could unlock a new form of contextual reasoning for discriminative models.

However, existing vision systems have no mechanism to determine which contextual information would actually improve their predictions.
The few methods that incorporate auxiliary inputs rely on fixed heuristics, \eg, visual similarity, class membership, or spatial proximity, rather than learning which examples are truly useful for the task. 
The success of retrieval-augmented generation (RAG) in language models~\cite{lewis2020retrieval, karpukhin2020dense} has inspired a broader family of retrieval-augmented learning (RAL) approaches~\cite{blattmann2022retrieval, hu2023reveal, li2024rag, yu2024visrag, faysse2024colpaliefficientdocumentretrieval, zheng2025retrieval}, where models retrieve external information to improve reasoning and prediction in generalist systems. 
However, for specialized models in visual recognition such as ViTs, retrieval remains a static preprocessing step: lacking mechanisms to process dynamic external context, they rely on pre-computed similarity metrics or embeddings such as CLIP or DINO~\cite{long2022retrieval, zhao2025retrieval, li2025rau}, implicitly assuming that perceptual similarity equates to usefulness.
This assumption is debatable, as visual similarity alone does not guarantee that an example will help a ViT make a better decision.

We propose {\oursfull{} (\ours)}, a framework that enables discriminative visual models to \textit{learn what helps}. 
\ours{} transforms context selection from a static, pre-defined operation into a learnable, task-aligned component of the visual pipeline. 
It comprises two jointly optimized modules: a \textit{selector network}, which learns to identify contextual examples that are most informative, and a \textit{downstream network}, which performs the primary vision task using both the input and the selected context (\cref{fig:introduction}). 
The selector is trained with a hybrid optimization scheme that combines gradient-based learning via differentiable relaxation with reinforcement learning for discrete selection, which allows it to receive feedback directly from the downstream objective and align selection with task performance. 

\ours{} enables discriminative models such as ViTs to learn which contextual information improves their predictions, introducing task-aligned retrieval for vision-only architectures. 
This is particularly valuable in data-limited domains like medical imaging, where paired data and annotations are scarce.
Our contributions are as follows:
\begin{enumerate}
    \item \textbf{Task-aligned context selection for ViTs.} 
    We present \oursfull{}, a framework that enables vision-only models to learn which auxiliary samples improve downstream performance.
    \item \textbf{A hybrid optimization strategy.} 
    We develop a hybrid training approach that combines gradient-based learning via differentiable relaxation with reinforcement learning for discrete selection, allowing the selector network to align its choices directly with downstream performance signals (\eg cross-entropy loss).
    \item \textbf{Comprehensive empirical validation.} 
    We evaluate \ours{} on $18$ datasets covering fine-grained natural image classification, medical image classification, and segmentation, demonstrating consistent improvements, especially in challenging or data-limited settings.
\end{enumerate}

\section{Related Work}
\label{sec:related}

Relevant prior work spans two main areas: retrieval-augmented learning in computer vision and optimization strategies for learnable sampling and discrete selection.

\vspace{3mm}
\noindent\textbf{Retrieval-Augmented Learning in Vision}

\noindent Retrieval-augmented learning enhances model performance by conditioning predictions on external information retrieved from large corpora. 
It originated in natural language processing through retrieval-augmented generation (RAG), where language models retrieve and condition on relevant text to solve knowledge-based tasks~\cite{pmlr-v119-guu20a, karpukhin2020dense, lewis2020retrieval, singh2021end, izacard2021distilling, asai2023selfrag, gao2024smartrag}. 
The idea has since expanded to multimodal reasoning~\cite{hu2023reveal, li2024rag, yu2024visrag, faysse2024colpaliefficientdocumentretrieval}, integrating visual and textual evidence, and to vision-only systems~\cite{long2022retrieval, blattmann2022retrieval, zhao2025retrieval} that enhance recognition or grounding by referencing retrieved images.

In these frameworks, retrieval is often treated as a latent variable~\cite{karpukhin2020dense, li2024rag}, and models are trained to identify embeddings that best support a generation or reasoning objective. 
This approach has proven effective for large generalist models that can dynamically condition on retrieved context. 
In computer vision, retrieval-augmented methods have been explored for long-tail recognition~\cite{long2022retrieval}, domain generalization~\cite{liu2023learning, liu2025few}, and few-shot learning~\cite{liu2025few, zhao2025retrieval}. 
REACT~\cite{liu2023learning} and SWAT~\cite{liu2025few} adapt vision-language models on retrieved web data, showing that retrieval can improve transfer when relevant examples are available. 
However, in both frameworks, retrieval is static and similarity-driven: retrieved images are chosen from CLIP or VLM embeddings and used only as additional training data. 
More broadly, retrieval in discriminative vision models remains a static preprocessing step, relying on pre-computed similarity metrics or frozen embeddings (\eg, CLIP or DINO)~\cite{long2022retrieval, zhao2025retrieval, li2025rau}, and implicitly assuming that perceptual similarity correlates with task usefulness. 
Unlike generalist systems that flexibly incorporate arbitrary external context, discriminative architectures such as ViTs are trained on fixed single-view inputs and lack mechanisms to dynamically integrate retrieved examples. 
In contrast, \ours{} redefines retrieval as part of the inference process itself, learning a per-instance, task-aligned selection policy that optimizes for downstream performance rather than resemblance. 

\vspace{3mm}
\noindent\textbf{Differentiable and Reinforcement-Based Selection}

\noindent Retrieval inherently involves discrete choices, posing challenges for end-to-end optimization. 
To address this, prior work has explored differentiable relaxations or reinforcement-based training that let gradients shape retrieval policies. 
Differentiable approximations include attention distillation~\cite{izacard2021distilling}, perplexity distillation~\cite{izacard2023atlas}, and attentive fusion~\cite{hu2023reveal}, which propagate task gradients through retrieval weights. 
Complementary reinforcement-based strategies explicitly optimize retrieval using task-dependent rewards~\cite{asai2023selfrag, li2024rag, gao2024smartrag, shi2025direct}. 
SmartRAG~\cite{gao2024smartrag} and RAG-DDR~\cite{li2024rag} introduced differentiable data rewards to jointly train retrieval and generation modules, while DRO~\cite{shi2025direct} treated retrieval as importance sampling over latent permutations, using policy-based reinforcement learning to maximize expected task performance. 
These works show that retrieval quality improves when selection is shaped by downstream outcomes rather than static metrics.

However, existing differentiable and reinforcement-based retrieval methods target generative or multimodal tasks. 
They have not been applied to discriminative visual settings, where the value of retrieved context must be measured by its contribution to predictive accuracy rather than text fluency or coherence. 
Our approach extends this principle to visual recognition: we introduce a hybrid framework that combines gradient-based embedding learning with policy-gradient optimization using downstream reward signals. 
This formulation enables task-aligned, learnable context selection for discriminative architectures such as ViTs, a capability not addressed by prior retrieval-augmented methods.

\begin{figure*}[t]
    \centering
    \includegraphics[width=0.95\textwidth]{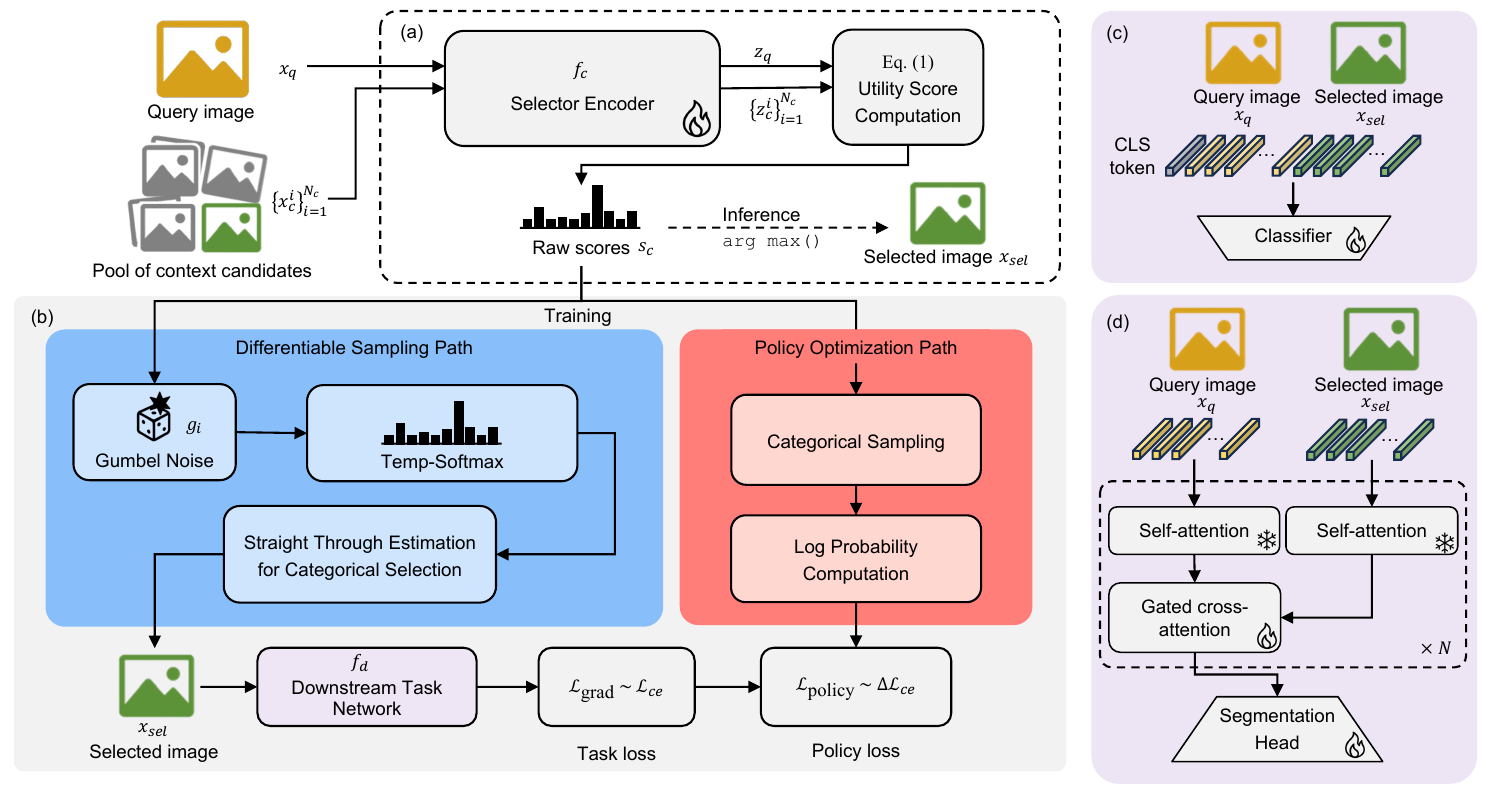}
    \caption{
    Architecture and training flow of Task-Aligned Context Selection (TACS).
    (a) The Selector processes a query image $x_q$ and candidate pool ${x_c^i}$ and selects the most helpful sample. 
    During training, the selection is optimized through two complementary optimization paths (b).
    The differentiable Sampling path (blue) uses the straight through Gumbel-Softmax reparameterization on the utility scores $s_c$ to implement a categorical selection of candidates, enabling end-to-end gradient flow via the task loss $\mathcal{L}_{\text{grad}}$.
    The policy path (red) samples a discrete action based on  $s_c$ and updates the Selector using policy gradients with a task-aligned reward $r(o,a)$ derived from downstream performance.
    Gradients and rewards jointly update shared Selector parameters.
    (c–d) Examples of downstream tasks: classification and segmentation. 
    The combined loss enables stable yet decisive task-aligned selection. 
    }
    \label{fig:method}
\end{figure*}

\section{Method}
\label{sec:method}

We propose \ours, a framework that enables discriminative vision models to learn which contextual examples most improve their predictions.
A \textit{Selector} retrieves informative samples from a candidate pool and is trained jointly with a \textit{Downstream Task Network} (\eg classifier or segmentation), making retrieval itself part of the learning objective.

Intuitively, the Selector learns a notion of \textit{helpfulness}: the expected gain in downstream task performance if a retrieved candidate is paired with the query.
But retrieval is inherently discrete, making end-to-end optimization difficult.
We address this with a {hybrid optimization strategy} combining two complementary signals:
(1) a \textit{differentiable sampling path} that provides stable gradient flow and models smooth utility relationships across candidates, and
(2) a \textit{policy optimization path} that refines selection through reward-based feedback from the downstream task,  trained jointly as illustrated in \cref{fig:method}.


\subsection{Architecture}

Given a query image \(x_q \in \mathbb{R}^{H \times W \times C}\) and a pool of \(N_c\) candidates \(\{x_c^i\}_{i=1}^{N_c}\),
the Selector backbone encodes them into \(z_q\) and \( z_c^i \in \mathbb{R}^{(L+1) \times D}\) where \(L\) denotes the number of spatial tokens and \(D\) is the feature size.
Each candidate’s \textit{utility score} is computed as
\begin{equation}
s_c^i = z_q^{\mathsf{T}} z_c^i ,
\end{equation}
and converted into selection probabilities
\begin{equation}
\label{eq:softmax}
p(x_c^i | x_q) =
\frac{\exp(s_c^i)}
{\sum_{j=1}^{N_c}\exp(s_c^j)} .
\end{equation}
During inference, the selected context is
\begin{equation}
x_{\text{sel}} = \arg\max_i p(x_c^i | x_q).
\end{equation}

The Downstream Task Network \(f_d\) predicts
\(\hat{y} = f_d(x_q, x_{\text{sel}})\),
and its gradients provide feedback to the Selector through both optimization paths (\cref{sec:hybrid}).
This joint training encourages embeddings that reflect \textit{task utility} rather than visual similarity, transforming retrieval from a static preprocessing step into a learnable component of a discriminative pipeline.

\subsection{Learning to Select via Hybrid Optimization}
\label{sec:hybrid}

Training the Selector involves discrete retrieval choices, which prevents direct gradient propagation.
To make retrieval learnable, we employ a {hybrid optimization strategy} that combines differentiable relaxation with policy-based reinforcement.
The differentiable path provides stable gradient flow and shapes smooth utility relationships between candidates, while the policy path reinforces discrete selection behaviors that truly improve downstream performance.

\paragraph{Differentiable sampling path.}
To enable end-to-end gradient flow, we adopt the straight-through Gumbel-Softmax estimator~\cite{DBLP:conf/iclr/JangGP17}, a differentiable approximation to categorical sampling.
This allows the Selector to generate one-hot selections during the forward pass while maintaining gradients during backpropagation.
The Gumbel-Softmax distribution provides a continuous relaxation of categorical distributions by adding Gumbel noise to logits:
\begin{equation}
    g_i = -\log(-\log(u_i)), \quad u_i \sim \mathcal{U}\left(0,1\right)
\end{equation}
The soft sample is then computed using a temperature-controlled softmax instead of \cref{eq:softmax}:
\begin{equation}
    p\left(x_c^i | x_q\right) = 
    \frac{\operatorname{exp}\left((s^i_c + g_i)/\tau\right)}
    {\sum_{j=1}^{N_c}\operatorname{exp}\left((s^j_c + g_j)/\tau\right)}
\end{equation} 
We apply straight through estimation on $p\left(x_c^i | x_q\right)$ for a categorical selection following~\cite{DBLP:conf/iclr/JangGP17}, to maintain gradient stability and selection sharpness.
The corresponding objective for the downstream task is
\begin{equation}
\mathcal{L}_{\text{grad}} =
\mathcal{L}_{ce}\!\left(f_d(x_q, x_{\text{sel}}), y\right),
\label{eq:hard_loss}
\end{equation}
where $\mathcal{L}_{ce}$ is the task loss.

\paragraph{Reward-based policy optimization path.}
While differentiable sampling stabilizes training, it does not explicitly evaluate whether adding a retrieved image actually improves the downstream prediction.
Ideally, a good selection should satisfy:
\begin{equation}
    \mathcal{L}_{ce}\!\left(f_d(x_q,x_c),y\right)
    < \mathcal{L}_{ce}\!\left(f_d(x_q,\emptyset),y\right),
\end{equation}
indicating that the retrieved context yields a lower task loss than using the query alone.
To capture this relative improvement, we optimize the Selector as a policy that receives reward feedback from the downstream model.
In this path, the Selector acts as an agent with observations $o=\{z_q, z_c^1, ..., z_c^{N_c}\}$, actions $a \in \{1, ..., N_c\}$, and a policy $\pi(a|o)=\operatorname{softmax}(\mathbf{s}(o))$ parameterized by the utility scores.
The reward measures the improvement in downstream performance:
\begin{equation}
r(o,a)=
\mathcal{L}_{ce}\!\left(f_d(x_q,\emptyset),y\right)
- \mathcal{L}_{ce}\!\left(f_d(x_q,x_c^{a}),y\right),
\end{equation}
where positive rewards indicate that the retrieved image improves task accuracy.
Gradients are detached from the Task model so that policy updates depend solely on reward signals.
The policy objective is then defined as:
\begin{equation}
\mathcal{L}_{\text{policy}} =
-\mathbb{E}_{a \sim \pi(\cdot|o)}
\!\big[\log \pi(a|o)\, A(o,a)\big],
\end{equation}
where \(A(o,a)\) denotes the advantage function, computed as the standardized reward within each batch to reduce gradient variance and stabilize updates.

\paragraph{Joint objective.}
Both optimization paths share the Selector parameters and are trained jointly, ensuring that gradients and rewards influence a shared notion of task utility.
The full training loss combines both components:
\begin{equation}
\mathcal{L}_{\text{TACS}} =
\mathcal{L}_{\text{grad}} +
\lambda\,\mathcal{L}_{\text{policy}},
\end{equation}
where $\lambda$ balances gradient-based and reward-based updates.

\section{Experiments}
\label{sec:experiments}

We evaluate \oursfull{} (\ours) on a diverse suite of visual recognition tasks to test three hypotheses:
(1)~\ours{} generalizes across tasks and domains, including natural and medical imaging;
(2)~its gains arise from learning task-aligned context selection; and
(3)~the learned selection policy captures interpretable contextual relationships that differ from conventional similarity-based retrieval.

\subsection{Datasets and Tasks}
\label{sec:datasets}

To assess generalization, we evaluate \ours{} across \textbf{18 datasets} spanning natural and medical imaging.  
These were chosen to represent complementary challenges: fine-grained recognition tests subtle inter-class boundaries, medical classification probes robustness under diagnostic ambiguity and data scarcity, and medical segmentation evaluates structured reasoning under limited supervision (\cref{tab:datasets}).

\vspace{3mm}
\noindent\textbf{Fine-grained natural image classification.}
We use 11 benchmarks including Aircraft \cite{maji2013fine},
Caltech-101 \cite{fei2004learning},
Caltech-256 \cite{griffin2007caltech},
CIFAR-10 \cite{krizhevsky2009learning},
CIFAR-100 \cite{krizhevsky2009learning},
CUB \cite{wah2011caltech},
DTD \cite{cimpoi2014describing},
Food-101 \cite{bossard2014food},
Oxford Pets \cite{parkhi2012cats},
Stanford Cars \cite{krause20133d},
and SUN397 \cite{xiao2010sun}. 
These datasets range from $5,000$ to $110,000$ images and emphasize subtle appearance variations and large intra-class diversity, an ideal setting to evaluate whether learned contextual pairing enhances discrimination.

\vspace{3mm}
\noindent\textbf{Medical image classification.}
We include four datasets from breast (DDSM \cite{lee2017curated}), skin (ISIC-2019 \cite{tschandl2018ham10000, codella2018skin, combalia2019bcn20000}), colorectal cancer (Col.~\cite{kather2016multi}) and retinal (APTOS2019 \cite{aptos2019-blindness-detection}) imaging.  
They feature small sample sizes and overlapping visual signatures, where helpful contextual examples may compensate for limited supervision.

\vspace{3mm}
\noindent\textbf{Medical segmentation.}
We further experiment on ISIC-2017 \cite{codella2018skin}, Kvasir-SEG \cite{jha2019kvasir}, and DRIVE \cite{staal2004ridge} to examine whether \ours{} can aid pixel-level predictions.  
These tasks evaluate if task-aligned retrieval supports spatial reasoning and fine boundary delineation.

\begin{table}[t]
\centering
\caption{Summary of datasets used in evaluation. The chosen datasets span diverse visual domains and supervision levels to test generalization and robustness.}
\label{tab:datasets}
\resizebox{\linewidth}{!}{
\begin{tabular}{lccc}
\toprule
Domain & Task & \#Datasets & Example Benchmarks \\
\midrule
Natural & Fine-grained classification & 11 & CUB, Cars, SUN397, ... \\
Medical & Classification & 4 & DDSM, ISIC, APTOS, ... \\
Medical & Segmentation & 3 & ISIC-2017, Kvasir, DRIVE \\
\bottomrule
\end{tabular}}
\end{table}

\vspace{3mm}
\noindent\textbf{Evaluation Metrics.}
Performance is reported using metrics standard to each task:
For fine-grained classification, we follow the evaluation protocols adopted in \cite{chen2020simple}, and report top-1 accuracy or mean per-class accuracy for each dataset.
For medical classification tasks, we report quadratic Cohen’s Kappa for
APTOS2019, accuracy for Colorectal, ROC-AUC for DDSM, and Recall on ISIC 2019, following \cite{huix2024natural}. 
For medical segmentation, we report Dice coefficient and IoU.

\subsection{Implementation Details}
\label{sec:implementation}

The Selector and Downstream Task Networks share a ViT-S/16 backbone.
All models are initialized with DINOv3 \cite{simeoni2025dinov3} pretrained weights and finetuned with AdamW optimizer \cite{loshchilov2017decoupled}. 
The implementation of the downstream multi-input model varies depending on the task.
For classification, we simply concatenate the patch embeddings of the main and the paired image before sending them to the transformer blocks and finetune the whole model;
For segmentation, we freeze the pretrained backbone and append a gated cross-attention block \cite{alayrac2022flamingo, lian2025describe} to each transformer block, to enhance the main image features, and adopt a lightweight DPT head \cite{ranftl2021vision} to obtain the final segmentation. 
We set Gumbel-Softmax temperature $\tau=0.1$ and hybrid weight $\lambda=0.5$ by default.  
Training uses 100 epochs with cosine annealing learning rate scheduling. 
Batch size is 64 for classification and 8 for segmentation. 
Images are resized to a resolution of $256 \times 256$, and augmented with horizontal and vertical flips and random resize crops during training.
All models are trained on a single NVIDIA A100 GPU.  
Each dataset follows its standard split, and results are averaged over three random seeds for robustness.

\paragraph{Candidate Pool Construction.}
For each dataset (except for DRIVE due to small size), we sample $20\%$ of the training images to form a fixed candidate pool used during training and evaluation. 
Query-associated images (e.g., from the same patient) are excluded across all paring operations to prevent data leakage.

\subsection{Baselines}
\label{sec:baselines}

We design baselines to progressively isolate the effect of task-aligned retrieval, ranging from no retrieval to fully learned selection:

\begin{itemize}
    \item \textbf{\nocontxt:} the model processes only the query image, without any contextual input.
    \item \textbf{\randctx:} a random candidate is paired with each query, representing retrieval without guidance.
    \item \textbf{\frozenretr:} retrieval based on precomputed DINOv3 embeddings \cite{zhao2025retrieval}; the retriever is frozen and non-learnable.
    \item \textbf{\featavg:} soft aggregation of the top-$k$ retrieved features into a continuous contextual representation according to \cref{eq:softmax} (no reinforcement).
\end{itemize}



\begin{table*}[t]
\caption{\textbf{Main results} on fine-grained and medical image classification benchmarks. 
The \textbf{best} results are highlighted for each task. 
} 
\centering
\small
\resizebox{\textwidth}{!}{%

\begin{tabular}{l@{\hspace{1mm}}ccccccccccccc@{\hspace{1mm}}cccccc}
    \toprule
    \adjustbox{angle=0}{\textbf{ }} & &
    \adjustbox{angle=0}{\textbf{Aircr.}} & 
    \adjustbox{angle=0}{\textbf{Cal101}} & 
    \adjustbox{angle=0}{\textbf{Cal256}} & 
    \adjustbox{angle=0}{\textbf{C10}} & 
    \adjustbox{angle=0}{\textbf{C100}} & 
    \adjustbox{angle=0}{\textbf{CUB}} & 
    \adjustbox{angle=0}{\textbf{DTD}} & 
    \adjustbox{angle=0}{\textbf{Food}} & 
    \adjustbox{angle=0}{\textbf{Pets}} & 
    \adjustbox{angle=0}{\textbf{Cars}} & 
    \adjustbox{angle=0}{\textbf{SUN}} &
    \adjustbox{angle=0}{\textbf{Avg.}} & & 
    \adjustbox{angle=0}{\textbf{APTOS}} & 
    \adjustbox{angle=0}{\textbf{Col.}} & 
    \adjustbox{angle=0}{\textbf{DDSM}} & 
    \adjustbox{angle=0}{\textbf{ISIC}} & 
    \adjustbox{angle=0}{\textbf{Avg.}} \\ 
    {} & &
    {mAcc $\uparrow$ } & 
    {mAcc $\uparrow$ } & 
    {mAcc $\uparrow$ } & 
    {Acc $\uparrow$ } & 
    {Acc $\uparrow$ } & 
    {Acc $\uparrow$ } & 
    {Acc $\uparrow$ } & 
    {Acc $\uparrow$ } & 
    {mAcc $\uparrow$ } & 
    {Acc $\uparrow$ } & 
    {Acc $\uparrow$ } &
    {- $\uparrow$} & & 
    {$\mathcal{K}$ $\uparrow$ } & 
    {Acc $\uparrow$ } & 
    {AUC $\uparrow$ } & 
    {Rec $\uparrow$ } & 
    {- $\uparrow$} \\ 
    \midrule
    \nocontxtsm   & {} & $84.3$ & $94.3$ & $87.8$ & $97.9$ & $87.8$ & $82.1$ & $76.1$ & $89.2$ & $89.8$ & $92.4$ & $68.0$ & $86.3$ & {} & $89.8$ & $96.2$ & $96.1$ & $82.0$ & $91.0$ \\
    \randctxsm    & {} & $82.8$ & $94.2$ & $86.3$ & $97.6$ & $87.6$ & $82.1$ & $74.6$ & $89.3$ & $90.0$ & $92.3$ & $69.0$ & $86.0$ & {} & $89.8$ & $96.8$ & $96.3$ & $83.5$ & $91.6$ \\
    \frozenretrsm    & {} & $85.1$ & $94.7$ & $87.3$ & $97.8$ & $88.4$ & $81.4$ & $75.4$ & $89.9$ & $90.3$ & $93.0$ & $69.2$ & $86.6$ & {} & $90.4$ & $97.0$ & $96.1$ & $84.5$ & $92.0$ \\
    \featavgsm.    & {} & $85.3$ & $94.3$ & $87.4$ & $97.8$ & $88.6$ & $82.6$ & $76.3$ & $90.5$ & $90.7$ & $93.2$ & $66.7$ & $86.7$ & {} & $90.2$ & $97.5$ & $96.6$ & $84.2$ & $92.1$ \\
    {\ours{}} (Ours)    & {} & $\mathbf{85.9}$ & $\mathbf{95.9}$ & $\mathbf{89.3}$ & $\mathbf{98.5}$ & $\mathbf{89.2}$ & $\mathbf{85.2}$ & $\mathbf{77.7}$ & $\mathbf{92.6}$ & $\mathbf{91.3}$ & $\mathbf{94.1}$ & $\mathbf{71.8}$ & $\mathbf{88.3}$ & {} & $\mathbf{91.1}$ & $\mathbf{97.9}$ & $\mathbf{97.4}$ & $\mathbf{85.2}$ & $\mathbf{92.9}$ \\
    \bottomrule
\end{tabular}
  
}%
\label{tab:main_cls}
\end{table*}

\subsection{Main Results}
\label{sec:main_results}

\cref{tab:main_cls} summarizes results across  fine-grained and medical classification datasets. 
\cref{tab:medical_seg} summarizes results on segmentation datasets.
\ours{} consistently outperforms all baselines across domains and tasks, validating Hypothesis~1 that task-aligned context selection generalizes widely.

\vspace{3mm}
\noindent\textbf{Fine-grained classification.}
Across 11 datasets, \ours{} improves accuracy by an average of {+1.7\%} over the frozen retrieval baseline, with gains up to +2.6\%~on SUN397 and +3.8\%~on CUB-200.  
The largest improvements occur on SUN397 and CUB-200, datasets characterized by high intra-class diversity and subtle inter-class boundaries. 
Here, similarity-based retrieval often reinforces redundancy by selecting near-duplicates (\eg, birds in similar poses or scenes under identical lighting), whereas \ours{} learns to retrieve complementary examples that expose discriminative contrasts such as different postures, lighting, or spatial layouts, helping the model refine fine-grained decision boundaries.

\begin{table}[t]
\caption{\textbf{Main results} on medical segmentation benchmarks. 
} 
\small
\begin{adjustbox}{width=0.99\columnwidth,center}
\begin{tabular}{lcccccc}
\toprule
\multirow{2}{*}{ } & 
\multicolumn{2}{c}{\textbf{DRIVE}} & 
\multicolumn{2}{c}{\textbf{ISIC 2017}} & 
\multicolumn{2}{c}{\textbf{Kvasir-SEG}} \\
    & Dice $\uparrow$       
    & IoU $\uparrow$      
    & Dice $\uparrow$       
    & IoU $\uparrow$      
    & Dice $\uparrow$       
    & IoU $\uparrow$ \\
\midrule
\nocontxtsm    & 76.2 & 64.6 & 83.3 & 74.4 & 84.6 & 77.0     \\
\randctxsm     & 76.7 & 64.9 & 83.7 & 76.3 & 85.6 & 79.9     \\
\frozenretrsm      & 77.5 & 64.8 & 84.4 & 76.0 & 86.3 & 80.0     \\ 
\featavgsm       & 77.6 & 64.8 & 84.5 & 75.6 & 86.4 & 79.8     \\
{\ours} (Ours)   & \textbf{78.1} & \textbf{65.4} & \textbf{85.0} & \textbf{76.5} & \textbf{87.1} & \textbf{81.1} \\
\bottomrule
\end{tabular}

\end{adjustbox}
\label{tab:medical_seg}
\end{table}

\vspace{3mm}
\noindent\textbf{Medical classification.}
Across Medical classification tasks, \ours{} consistently outperforms all baselines, achieving an average gain of {0.9\%} over DINO similarity retrieval.
The largest gain occurs on DDSM, where the learned selection policy provides {+1.3\%} AUC improvement.

\vspace{3mm}
\noindent\textbf{Medical segmentation.}
Across DRIVE, ISIC-2017, and Kvasir-SEG, \ours{} consistently improves segmentation quality over all baselines, achieving up to {+1.1 IoU} on Kvasir-SEG.
These gains reflect the model’s ability to retrieve contextually informative examples that empirically benefit dense prediction tasks. 

\subsection{Ablation Studies }
\label{sec:ablation}

To test Hypothesis~2 (H2), that performance gains stem from learned, task-aligned retrieval, we conduct ablation experiments to isolate the effects of (a) contextual pairing and (b) the hybrid optimization components.

\begin{table*}[t]
\caption{Ablations on methods for constructing context pairs. The context image is not provided, provided as a blank image, a duplicate of the query image, or a noisy version of the query image.} 
\centering
\small
\begin{adjustbox}{width=0.99\textwidth,center}

\begin{tabular}{lcccccccccccccccc}
\toprule
 & \textbf{Aircr.} & \textbf{Cal101} & \textbf{Cal256} & \textbf{C10} & \textbf{C100} & \textbf{CUB} & \textbf{DTD} & \textbf{Food} & \textbf{Pets} & \textbf{Cars} & \textbf{SUN} & \textbf{APTOS} & \textbf{Col.} & \textbf{DDSM} & \textbf{ISIC} & \textbf{Avg.} \\
{\textbf{Context Image}} &
{mAcc $\uparrow$ } & 
    {mAcc $\uparrow$ } & 
    {mAcc $\uparrow$ } & 
    {Acc $\uparrow$ } & 
    {Acc $\uparrow$ } & 
    {Acc $\uparrow$ } & 
    {Acc $\uparrow$ } & 
    {Acc $\uparrow$ } & 
    {mAcc $\uparrow$ } & 
    {Acc $\uparrow$ } & 
    {Acc $\uparrow$ } &
    {$\mathcal{K}$ $\uparrow$ } & 
    {Acc $\uparrow$ } & 
    {AUC $\uparrow$ } & 
    {Rec $\uparrow$ } & 
    {- $\uparrow$} \\
\midrule
{No context image}   & 84.3 & 94.3 & 87.8 & 97.9 & 87.8 & 82.1 & 76.1 & 89.2 & 89.8 & 92.4 & 68.0 & 89.8 & 96.2 & 96.1 & 82.0 & 87.6 \\
{Blank Image}    & 84.2 & 94.1 & 87.3 & 97.7 & 87.6 & 82.4 & 74.7 & 89.1 & 89.4 & 92.2 & 67.6 & 90.0 & 96.7 & 96.2 & 83.7 & 87.5 \\
{Duplicate Query}& 84.1 & 94.1 & 87.0 & 97.9 & 87.4 & 82.6 & 75.9 & 89.3 & 89.9 & 92.4 & 67.5 & 89.9 & 96.6 & 96.4 & 83.9 & 87.7 \\
{Noisy Query}    & 84.4 & 94.4 & 87.2 & 97.6 & 88.1 & 82.6 & 75.6 & 89.3 & 89.6 & 92.5 & 67.8 & 89.8 & 96.8 & 96.5 & 84.2 & 87.8 \\
{Similar image (DINOv3)}& 85.1 & 94.7 & 87.3 & 97.8 & 88.4 & 81.4 & 75.4 & 89.9 & 90.3 & 93.0 & 69.2 & 90.4 & 97.0 & 96.1 & 84.5 & 88.0 \\
{\ours{}} Context image & \textbf{85.9} & \textbf{95.9} & \textbf{89.3} & \textbf{98.5} & \textbf{89.2} & \textbf{85.2} & \textbf{77.7} & \textbf{92.6} & \textbf{91.3} & \textbf{94.1} & \textbf{71.8} & \textbf{91.1} & \textbf{97.9} & \textbf{97.4} & \textbf{85.2} & \textbf{89.5} \\
\bottomrule
\end{tabular}

\end{adjustbox}
\label{tab:ablation_pair}
\end{table*}

\paragraph{Effect of contextual pairing.}
We compare \ours{} with several control settings that share identical architectures but differ in how contextual images are paired. 
Specifically, we replace learned retrieval with 
(i)~no context image.
(ii)~an empty image, 
(iii)~a duplicate of the query image, 
(iv)~a noisy version of the query image, and 
(v)~a static similarity-based retrieval using frozen DINOv3 embeddings (as in the main results).
Results appear in \cref{tab:ablation_pair}.
Across both natural and medical datasets, these alternatives produce smaller or inconsistent gains, demonstrating that improvements do not arise from simply concatenating more image tokens or from model capacity alone.
Instead, performance scales with the \textit{informativeness} of the retrieved image.
Providing an empty context image or duplicating the query image as the context image has the same effect as not providing a context image at all, a loss of approximately 2\% performance. Fixed similar image retrieval improves performance somewhat, but significantly less than task-aligned context selection, where examples are empirically proven more useful for the downstream task.

\begin{table}[t]
\caption{Ablations on components of \ours{}.} 
\centering
\small
\begin{adjustbox}{width=0.99\columnwidth,center}

\begin{tabular}{lcc}
\toprule
Method              & Fine-Grained & Med Cls.   \\ 
\midrule
{\frozenretrsm }        & 86.6         & 92.0   \\
{Differentiable selection (no policy)}         & 86.7         & 92.1   \\
{Policy-based selection (no Gumbal softmax)}         & 87.2         & 92.4   \\
Full dual optimization (\ours{})  & 88.3         & 92.9   \\
\bottomrule
\end{tabular}
\end{adjustbox}
\label{tab:ablation_component}
\end{table}

\paragraph{Effect of optimization components.}
To assess the role of each optimization component, we ablate the differentiable and policy-based objectives of \ours{}.
We test four variants:
(a)~a fixed retriever with frozen embeddings,
(b)~soft (differentiable) selection only,
(c)~hard (policy-based) selection only, and
(d)~the full dual-path model.
Results in \cref{tab:ablation_component} show that while each path individually improves over static retrieval, their combination yields the most stable and highest performance across datasets.
The differentiable sampling path provides smooth supervision during early training, while the policy optimization path refines discrete selection decisions that maximize downstream reward.
This confirms H2: performance gains stem from the learned, task-aligned selection process driven jointly by smooth gradient supervision and discrete reward feedback.

\subsection{Analysis and Interpretability}
\label{sec:interpretability}

Finally, we investigate Hypothesis~3 (H3): that the learned selection policy captures interpretable contextual relationships distinct from conventional similarity retrieval.

\begin{figure}[t]
    \centering
    \begin{subfigure}[t]{0.45\linewidth}
        \centering
        \includegraphics[width=\columnwidth]{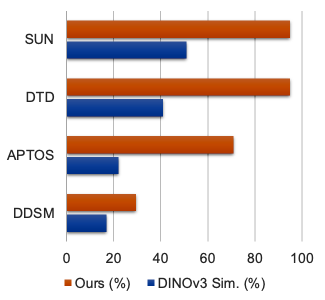}
        \caption{Cross-class Selection Rate}
    \end{subfigure}%
    ~ 
    \begin{subfigure}[t]{0.45\linewidth}
        \centering
        \includegraphics[width=\columnwidth]{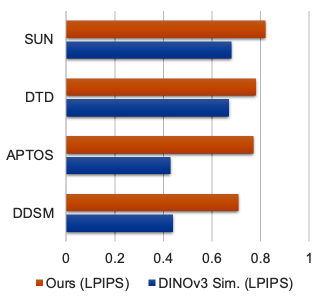}
        \caption{LPIPS Loss}
    \end{subfigure}
    \caption{\textbf{Statistical analysis of selected context pairs.}
Comparison of cross-class selection rates and mean LPIPS distances for pairs chosen by different retrieval strategies.
\ours{} consistently selects a higher proportion of cross-class and perceptually diverse (higher LPIPS) candidates than a fixed retriever, particularly on complex datasets such as DTD and SUN397.
This indicates that the learned policy favors \emph{complementary} rather than merely \emph{similar} examples, enabling richer contextual reasoning.
}
    \label{fig:statistical}
\end{figure}

\paragraph{Analysis of selection behavior.}
We analyze the distribution of selected pairs across class boundaries and visual similarity.
\cref{fig:statistical} compares the class-overlap rate and LPIPS distance between the query and the retrieved example for DINO similarity and \ours{}.
While similarity-based methods overwhelmingly select near-duplicates, \ours{} increases cross-class selection rates by 40--70\% and retrieves examples with higher perceptual diversity.
For instance, on the APTOS retinal dataset, \ours{} often pairs mild and severe diabetic retinopathy cases, enabling the model to contrast lesion severity; on SUN397, it retrieves scenes in contrasting environments to clarify fine-grained boundaries.
These findings indicate that the learned policy captures task-relevant complementarity rather than redundancy.
LPIPS tells a similar story, where across all datasets retrieved images from \ours{} display less visual similarity to the query image than those retrieved according to DINO pretrained features. 

\begin{figure*}[t]
    \centering
    \includegraphics[width=\textwidth]{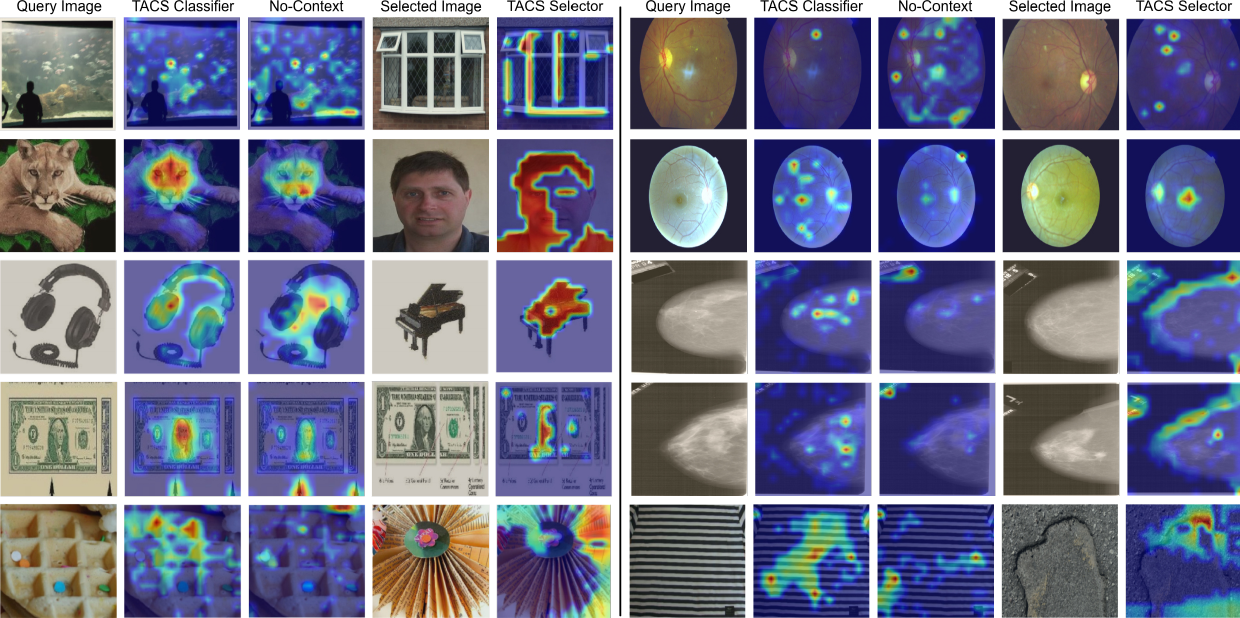}
    \caption{
    For each example, we show (1) the query image, (2) the \ours{} classifier’s attention on the query image (which also receives the selected context image), (3) the no-context classifier’s attention on the same image, (4) the context image selected by the Selector, and (5) the Selector’s attention on the selected image.
\ours{} consistently focuses the classifier on more discriminative regions than the no-context model. The Selector attends to broad global structures, often choosing complementary rather than visually similar images (\eg, retrieving a piano to help disambiguate headphones), enabling the classifier to refine fine-grained decisions through contrasting contextual cues.
    }
    \label{fig:qualitative}
\end{figure*}

\paragraph{Qualitative visualization.}
\cref{fig:qualitative} illustrates qualitative examples comparing similarity-based and task-aligned retrieval across medical and natural domains.
A consistent pattern emerges: the Selector and Task networks develop distinct yet complementary attention behaviors.
The Selector focuses on broad structural or contextual patterns such as vessel topology in retinal scans, breast structure in mammography, or geometric layout in scenes, while the Task network refines its attention to fine-grained, discriminative details within the query image.
In medical datasets, the retrieved examples often highlight diagnostically relevant contrasts, such as a mass boundary or vascular abnormality that helps the classifier localize pathology more precisely.
In fine-grained recognition, \ours{} retrieves semantically related but visually contrasting images that clarify decision boundaries.
For example, in SUN397, when the query depicts an aquarium, \ours{} retrieves a window façade—an image that shares planar reflections and glass textures yet exhibits the rigid framing and outdoor illumination absent in the query.
This counterfactual pairing allows the model to reason contrastively, reinforcing its aquarium prediction by recognizing what distinguishing features are \emph{missing}.
Similar cases are observed on Caltech-101 and DTD as well.
Notably, \ours{} also selects similar samples when needed, such as identifying the dollar object.
Overall, these results suggest that \ours{} learns to retrieve complementary, helpful examples, enabling the system to reason through structural analogy and contrast adaptively instead of mere similarity.

\section{Discussion}
\label{sec:discussion}

\paragraph{H1: Generalization across tasks and domains.}
Across 18 datasets spanning natural and medical imaging, \ours{} consistently improves performance over all baselines, validating its generality. 
The strongest gains appear in domains with high intra-class ambiguity and limited supervision, such as CUB-200, SUN397, and DDSM. 
These results show that task-aligned retrieval is not domain-specific: the same mechanism that helps a model distinguish sceneries can also improve diagnostic precision in medical imaging. 
This demonstrates that learning \emph{what helps} is a transferable inductive bias that complements diverse  tasks.

\paragraph{H2: Gains stem from learned retrieval.}
Ablation studies confirm that improvements arise from the task-aligned selection process itself rather than from additional parameters or data exposure. 
Both differentiable and policy-based optimization paths contribute complementary strengths: the former stabilizes early training by shaping continuous utility relationships, while the latter sharpens discrete decision boundaries through reward-based feedback. 
Together, these components enable \ours{} to select examples that meaningfully improve downstream accuracy—a capability static similarity-based methods lack.

\paragraph{H3: Interpretable contextual reasoning.}
Visualizations of selector and classifier attention (\cref{fig:qualitative}) reveal a learned division of labor: the Selector emphasizes global structure and contextual cues, while the Task model focuses on fine discriminative regions. 
In several cases, \ours{} resolves ambiguity by retrieving cross-class examples that clarify decision boundaries—for instance, retrieving a window image when classifying an aquarium, helping the model recognize the absence of window frames and confirm the aquarium label. 
This behavior suggests that \ours{} develops an implicit understanding of how contextual contrasts enhance recognition, offering a concrete form of interpretability grounded in task benefit rather than visual similarity.

\paragraph{Broader implications.}
Task-aligned retrieval reframes how discriminative models use contextual data. 
Rather than treating retrieval as a preprocessing step, \ours{} integrates it into the reasoning process itself, allowing the model to dynamically identify which information improves its decisions. 
This insight connects to broader topics such as data curation, active learning, and visual reasoning, suggesting that learning to select \emph{helpful} examples may serve as a unifying principle for adaptive visual intelligence.

\paragraph{Limitations and future work.}
Current limitations include the reliance on a fixed candidate pool and the use of a single retrieved example per query. 
Applications to video, detection, or multimodal decision-making represent promising next steps.

\section{Conclusion}
\label{sec:conclusion}

We introduced \ours{}, a framework that enables discriminative vision models to learn what helps.
By jointly optimizing a Selector and Task network through a hybrid differentiable–policy objective, \ours{} transforms context retrieval from a static similarity heuristic into a task-aligned inference module.
Experiments across 18 datasets show consistent gains, especially in high-ambiguity and data-limited settings, and qualitative analyses reveal complementary retrieval strategies that differ from nearest-neighbor baselines.
This work opens a new direction for retrieval-augmented vision: using selective external context as a learnable reasoning mechanism rather than a preprocessing step. Future work includes scaling to larger pools and extending task-aligned retrieval to broader vision reasoning pipelines.

\paragraph{\textbf{Acknowledgments.}}
This work was supported by the Wallenberg AI, Autonomous Systems and Software Program (WASP).
We acknowledge the Berzelius computational resources provided by the Knut and Alice Wallenberg Foundation at the National Supercomputer Centre and the computational resources provided by the National Academic Infrastructure for Supercomputing in Sweden (NAISS).
{
    \small
    \bibliographystyle{ieeenat_fullname}
    \bibliography{main}

\begin{thebibliography}{51}
\providecommand{\natexlab}[1]{#1}
\providecommand{\url}[1]{\texttt{#1}}
\expandafter\ifx\csname urlstyle\endcsname\relax
  \providecommand{\doi}[1]{doi: #1}\else
  \providecommand{\doi}{doi: \begingroup \urlstyle{rm}\Url}\fi

\bibitem[Alayrac et~al.(2022)Alayrac, Donahue, Luc, Miech, Barr, Hasson, Lenc, Mensch, Millican, Reynolds, et~al.]{alayrac2022flamingo}
Jean-Baptiste Alayrac, Jeff Donahue, Pauline Luc, Antoine Miech, Iain Barr, Yana Hasson, Karel Lenc, Arthur Mensch, Katherine Millican, Malcolm Reynolds, et~al.
\newblock Flamingo: a visual language model for few-shot learning.
\newblock \emph{Advances in neural information processing systems}, 35:\penalty0 23716--23736, 2022.

\bibitem[Alzahrani et~al.(2024)Alzahrani, Usman, Anwar, and Helmy]{alzahrani2024selective}
Mona Alzahrani, Muhammad Usman, Saeed Anwar, and Tarek Helmy.
\newblock Selective multi-view deep model for 3d object classification.
\newblock In \emph{Proceedings of the ieee/cvf conference on computer vision and pattern recognition}, pages 728--736, 2024.

\bibitem[Asai et~al.(2023)Asai, Wu, Wang, Sil, and Hajishirzi]{asai2023selfrag}
Akari Asai, Zeqiu Wu, Yizhong Wang, Avirup Sil, and Hannaneh Hajishirzi.
\newblock {Self-RAG}: Learning to retrieve, generate, and critique through self-reflection.
\newblock \emph{arXiv preprint arXiv:2310.11511}, 2023.

\bibitem[Blattmann et~al.(2022)Blattmann, Rombach, Oktay, M{\"u}ller, and Ommer]{blattmann2022retrieval}
Andreas Blattmann, Robin Rombach, Kaan Oktay, Jonas M{\"u}ller, and Bj{\"o}rn Ommer.
\newblock Retrieval-augmented diffusion models.
\newblock \emph{Advances in Neural Information Processing Systems}, 35:\penalty0 15309--15324, 2022.

\bibitem[Bolya et~al.(2025)Bolya, Huang, Sun, Cho, Madotto, Wei, Ma, Zhi, Rajasegaran, Rasheed, Wang, Monteiro, Xu, Dong, Ravi, Li, Doll{\'a}r, and Feichtenhofer]{bolya2025PerceptionEncoder}
Daniel Bolya, Po-Yao Huang, Peize Sun, Jang~Hyun Cho, Andrea Madotto, Chen Wei, Tengyu Ma, Jiale Zhi, Jathushan Rajasegaran, Hanoona Rasheed, Junke Wang, Marco Monteiro, Hu Xu, Shiyu Dong, Nikhila Ravi, Daniel Li, Piotr Doll{\'a}r, and Christoph Feichtenhofer.
\newblock Perception encoder: The best visual embeddings are not at the output of the network.
\newblock \emph{arXiv:2504.13181}, 2025.

\bibitem[Bossard et~al.(2014)Bossard, Guillaumin, and Van~Gool]{bossard2014food}
Lukas Bossard, Matthieu Guillaumin, and Luc Van~Gool.
\newblock Food-101--mining discriminative components with random forests.
\newblock In \emph{European conference on computer vision}, pages 446--461. Springer, 2014.

\bibitem[Chen et~al.(2020)Chen, Kornblith, Norouzi, and Hinton]{chen2020simple}
Ting Chen, Simon Kornblith, Mohammad Norouzi, and Geoffrey Hinton.
\newblock A simple framework for contrastive learning of visual representations.
\newblock In \emph{International conference on machine learning}, pages 1597--1607. PmLR, 2020.

\bibitem[Cimpoi et~al.(2014)Cimpoi, Maji, Kokkinos, Mohamed, and Vedaldi]{cimpoi2014describing}
Mircea Cimpoi, Subhransu Maji, Iasonas Kokkinos, Sammy Mohamed, and Andrea Vedaldi.
\newblock Describing textures in the wild.
\newblock In \emph{Proceedings of the IEEE conference on computer vision and pattern recognition}, pages 3606--3613, 2014.

\bibitem[Codella et~al.(2018)Codella, Gutman, Celebi, Helba, Marchetti, Dusza, Kalloo, Liopyris, Mishra, Kittler, et~al.]{codella2018skin}
Noel~CF Codella, David Gutman, M~Emre Celebi, Brian Helba, Michael~A Marchetti, Stephen~W Dusza, Aadi Kalloo, Konstantinos Liopyris, Nabin Mishra, Harald Kittler, et~al.
\newblock Skin lesion analysis toward melanoma detection: A challenge at the 2017 international symposium on biomedical imaging (isbi), hosted by the international skin imaging collaboration (isic).
\newblock In \emph{2018 IEEE 15th international symposium on biomedical imaging (ISBI 2018)}, pages 168--172. IEEE, 2018.

\bibitem[Combalia et~al.(2019)Combalia, Codella, Rotemberg, Helba, Vilaplana, Reiter, Carrera, Barreiro, Halpern, Puig, et~al.]{combalia2019bcn20000}
Marc Combalia, Noel~CF Codella, Veronica Rotemberg, Brian Helba, Veronica Vilaplana, Ofer Reiter, Cristina Carrera, Alicia Barreiro, Allan~C Halpern, Susana Puig, et~al.
\newblock Bcn20000: Dermoscopic lesions in the wild.
\newblock \emph{arXiv preprint arXiv:1908.02288}, 2019.

\bibitem[Faysse et~al.(2024)Faysse, Sibille, Wu, Omrani, Viaud, Hudelot, and Colombo]{faysse2024colpaliefficientdocumentretrieval}
Manuel Faysse, Hugues Sibille, Tony Wu, Bilel Omrani, Gautier Viaud, Céline Hudelot, and Pierre Colombo.
\newblock Colpali: Efficient document retrieval with vision language models, 2024.

\bibitem[Fei-Fei et~al.(2004)Fei-Fei, Fergus, and Perona]{fei2004learning}
Li Fei-Fei, Rob Fergus, and Pietro Perona.
\newblock Learning generative visual models from few training examples: An incremental bayesian approach tested on 101 object categories.
\newblock In \emph{2004 conference on computer vision and pattern recognition workshop}, pages 178--178. IEEE, 2004.

\bibitem[Gao et~al.(2024)Gao, Li, Li, Fu, and Dai]{gao2024smartrag}
Jingsheng Gao, Linxu Li, Weiyuan Li, Yuzhuo Fu, and Bin Dai.
\newblock Smartrag: Jointly learn rag-related tasks from the environment feedback.
\newblock \emph{arXiv preprint arXiv:2410.18141}, 2024.

\bibitem[Griffin et~al.(2007)Griffin, Holub, Perona, et~al.]{griffin2007caltech}
Gregory Griffin, Alex Holub, Pietro Perona, et~al.
\newblock Caltech-256 object category dataset.
\newblock Technical report, Technical Report 7694, California Institute of Technology Pasadena, 2007.

\bibitem[Guu et~al.(2020)Guu, Lee, Tung, Pasupat, and Chang]{pmlr-v119-guu20a}
Kelvin Guu, Kenton Lee, Zora Tung, Panupong Pasupat, and Mingwei Chang.
\newblock Retrieval augmented language model pre-training.
\newblock In \emph{Proceedings of the 37th International Conference on Machine Learning}, pages 3929--3938. PMLR, 2020.

\bibitem[Hou et~al.(2024)Hou, Gould, and Zheng]{hou2024learning}
Yunzhong Hou, Stephen Gould, and Liang Zheng.
\newblock Learning to select views for efficient multi-view understanding.
\newblock In \emph{Proceedings of the IEEE/CVF Conference on Computer Vision and Pattern Recognition}, pages 20135--20144, 2024.

\bibitem[Hu et~al.(2023)Hu, Iscen, Sun, Wang, Chang, Sun, Schmid, Ross, and Fathi]{hu2023reveal}
Ziniu Hu, Ahmet Iscen, Chen Sun, Zirui Wang, Kai-Wei Chang, Yizhou Sun, Cordelia Schmid, David~A Ross, and Alireza Fathi.
\newblock Reveal: Retrieval-augmented visual-language pre-training with multi-source multimodal knowledge memory.
\newblock In \emph{Proceedings of the IEEE/CVF conference on computer vision and pattern recognition}, pages 23369--23379, 2023.

\bibitem[Huix et~al.(2024)Huix, Ganeshan, Haslum, S{\"o}derberg, Matsoukas, and Smith]{huix2024natural}
Joana~Pal{\'e}s Huix, Adithya~Raju Ganeshan, Johan~Fredin Haslum, Magnus S{\"o}derberg, Christos Matsoukas, and Kevin Smith.
\newblock Are natural domain foundation models useful for medical image classification?
\newblock In \emph{Proceedings of the IEEE/CVF winter conference on applications of computer vision}, pages 7634--7643, 2024.

\bibitem[Izacard and Grave(2021)]{izacard2021distilling}
Gautier Izacard and Edouard Grave.
\newblock Distilling knowledge from reader to retriever for question answering.
\newblock In \emph{International Conference on Learning Representations}, 2021.

\bibitem[Izacard et~al.(2023)Izacard, Lewis, Lomeli, Hosseini, Petroni, Schick, Dwivedi-Yu, Joulin, Riedel, and Grave]{izacard2023atlas}
Gautier Izacard, Patrick Lewis, Maria Lomeli, Lucas Hosseini, Fabio Petroni, Timo Schick, Jane Dwivedi-Yu, Armand Joulin, Sebastian Riedel, and Edouard Grave.
\newblock Atlas: Few-shot learning with retrieval augmented language models.
\newblock \emph{Journal of Machine Learning Research}, 24\penalty0 (251):\penalty0 1--43, 2023.

\bibitem[Jang et~al.(2017)Jang, Gu, and Poole]{DBLP:conf/iclr/JangGP17}
Eric Jang, Shixiang Gu, and Ben Poole.
\newblock Categorical reparameterization with gumbel-softmax.
\newblock In \emph{5th International Conference on Learning Representations, {ICLR} 2017, Toulon, France, April 24-26, 2017, Conference Track Proceedings}. OpenReview.net, 2017.

\bibitem[Jha et~al.(2019)Jha, Smedsrud, Riegler, Halvorsen, De~Lange, Johansen, and Johansen]{jha2019kvasir}
Debesh Jha, Pia~H Smedsrud, Michael~A Riegler, P{\aa}l Halvorsen, Thomas De~Lange, Dag Johansen, and H{\aa}vard~D Johansen.
\newblock Kvasir-seg: A segmented polyp dataset.
\newblock In \emph{International conference on multimedia modeling}, pages 451--462. Springer, 2019.

\bibitem[Karpukhin et~al.(2020)Karpukhin, Oguz, Min, Lewis, Wu, Edunov, Chen, and Yih]{karpukhin2020dense}
Vladimir Karpukhin, Barlas Oguz, Sewon Min, Patrick~SH Lewis, Ledell Wu, Sergey Edunov, Danqi Chen, and Wen-tau Yih.
\newblock Dense passage retrieval for open-domain question answering.
\newblock In \emph{EMNLP (1)}, pages 6769--6781, 2020.

\bibitem[Karthik et~al.(2019)Karthik, Maggie, and Dane]{aptos2019-blindness-detection}
Karthik, Maggie, and Sohier Dane.
\newblock Aptos 2019 blindness detection.
\newblock \url{https://kaggle.com/competitions/aptos2019-blindness-detection}, 2019.
\newblock Kaggle.

\bibitem[Kather et~al.(2016)Kather, Weis, Bianconi, Melchers, Schad, Gaiser, Marx, and Z{\"o}llner]{kather2016multi}
Jakob~Nikolas Kather, Cleo-Aron Weis, Francesco Bianconi, Susanne~M Melchers, Lothar~R Schad, Timo Gaiser, Alexander Marx, and Frank~Gerrit Z{\"o}llner.
\newblock Multi-class texture analysis in colorectal cancer histology.
\newblock \emph{Scientific reports}, 6\penalty0 (1):\penalty0 1--11, 2016.

\bibitem[Krause et~al.(2013)Krause, Stark, Deng, and Fei-Fei]{krause20133d}
Jonathan Krause, Michael Stark, Jia Deng, and Li Fei-Fei.
\newblock 3d object representations for fine-grained categorization.
\newblock In \emph{Proceedings of the IEEE international conference on computer vision workshops}, pages 554--561, 2013.

\bibitem[Krizhevsky et~al.(2009)Krizhevsky, Hinton, et~al.]{krizhevsky2009learning}
Alex Krizhevsky, Geoffrey Hinton, et~al.
\newblock Learning multiple layers of features from tiny images.
\newblock 2009.

\bibitem[Lee et~al.(2017)Lee, Gimenez, Hoogi, Miyake, Gorovoy, and Rubin]{lee2017curated}
Rebecca~Sawyer Lee, Francisco Gimenez, Assaf Hoogi, Kanae~Kawai Miyake, Mia Gorovoy, and Daniel~L Rubin.
\newblock A curated mammography data set for use in computer-aided detection and diagnosis research.
\newblock \emph{Scientific data}, 4\penalty0 (1):\penalty0 1--9, 2017.

\bibitem[Lewis et~al.(2020)Lewis, Perez, Piktus, Petroni, Karpukhin, Goyal, K{\"u}ttler, Lewis, Yih, Rockt{\"a}schel, et~al.]{lewis2020retrieval}
Patrick Lewis, Ethan Perez, Aleksandra Piktus, Fabio Petroni, Vladimir Karpukhin, Naman Goyal, Heinrich K{\"u}ttler, Mike Lewis, Wen-tau Yih, Tim Rockt{\"a}schel, et~al.
\newblock Retrieval-augmented generation for knowledge-intensive nlp tasks.
\newblock \emph{Advances in neural information processing systems}, 33:\penalty0 9459--9474, 2020.

\bibitem[Li et~al.(2024)Li, Mei, Liu, Yan, Wang, Yu, Zeng, Chen, Yu, Liu, et~al.]{li2024rag}
Xinze Li, Sen Mei, Zhenghao Liu, Yukun Yan, Shuo Wang, Shi Yu, Zheni Zeng, Hao Chen, Ge Yu, Zhiyuan Liu, et~al.
\newblock Rag-ddr: Optimizing retrieval-augmented generation using differentiable data rewards.
\newblock \emph{arXiv preprint arXiv:2410.13509}, 2024.

\bibitem[Li et~al.(2025)Li, Liu, Guo, Zhao, Zhang, Chen, Mailhe, Mukherjee, Chen, and Sun]{li2025rau}
Yiwei Li, Yikang Liu, Jiaqi Guo, Lin Zhao, Zheyuan Zhang, Xiao Chen, Boris Mailhe, Ankush Mukherjee, Terrence Chen, and Shanhui Sun.
\newblock Rau: Reference-based anatomical understanding with vision language models.
\newblock \emph{arXiv preprint arXiv:2509.22404}, 2025.

\bibitem[Lian et~al.(2025)Lian, Ding, Ge, Liu, Mao, Li, Pavone, Liu, Darrell, Yala, et~al.]{lian2025describe}
Long Lian, Yifan Ding, Yunhao Ge, Sifei Liu, Hanzi Mao, Boyi Li, Marco Pavone, Ming-Yu Liu, Trevor Darrell, Adam Yala, et~al.
\newblock Describe anything: Detailed localized image and video captioning.
\newblock \emph{arXiv preprint arXiv:2504.16072}, 2025.

\bibitem[Liu et~al.(2023)Liu, Son, Yang, Liu, Gao, Lee, and Li]{liu2023learning}
Haotian Liu, Kilho Son, Jianwei Yang, Ce Liu, Jianfeng Gao, Yong~Jae Lee, and Chunyuan Li.
\newblock Learning customized visual models with retrieval-augmented knowledge.
\newblock In \emph{Proceedings of the IEEE/CVF Conference on Computer Vision and Pattern Recognition}, pages 15148--15158, 2023.

\bibitem[Liu et~al.(2025)Liu, Zhang, Parashar, and Kong]{liu2025few}
Tian Liu, Huixin Zhang, Shubham Parashar, and Shu Kong.
\newblock Few-shot recognition via stage-wise retrieval-augmented finetuning.
\newblock In \emph{Proceedings of the Computer Vision and Pattern Recognition Conference}, pages 15086--15097, 2025.

\bibitem[Long et~al.(2022)Long, Yin, Ajanthan, Nguyen, Purkait, Garg, Blair, Shen, and Van~den Hengel]{long2022retrieval}
Alexander Long, Wei Yin, Thalaiyasingam Ajanthan, Vu Nguyen, Pulak Purkait, Ravi Garg, Alan Blair, Chunhua Shen, and Anton Van~den Hengel.
\newblock Retrieval augmented classification for long-tail visual recognition.
\newblock In \emph{Proceedings of the IEEE/CVF conference on computer vision and pattern recognition}, pages 6959--6969, 2022.

\bibitem[Loshchilov and Hutter(2017)]{loshchilov2017decoupled}
Ilya Loshchilov and Frank Hutter.
\newblock Decoupled weight decay regularization.
\newblock \emph{arXiv preprint arXiv:1711.05101}, 2017.

\bibitem[Maji et~al.(2013)Maji, Rahtu, Kannala, Blaschko, and Vedaldi]{maji2013fine}
Subhransu Maji, Esa Rahtu, Juho Kannala, Matthew Blaschko, and Andrea Vedaldi.
\newblock Fine-grained visual classification of aircraft.
\newblock \emph{arXiv preprint arXiv:1306.5151}, 2013.

\bibitem[Nath et~al.(2025)Nath, Li, Yang, Myronenko, Zheng, Lu, Liu, Yin, Law, Tang, et~al.]{nath2025vila}
Vishwesh Nath, Wenqi Li, Dong Yang, Andriy Myronenko, Mingxin Zheng, Yao Lu, Zhijian Liu, Hongxu Yin, Yee~Man Law, Yucheng Tang, et~al.
\newblock Vila-m3: Enhancing vision-language models with medical expert knowledge.
\newblock In \emph{Proceedings of the Computer Vision and Pattern Recognition Conference}, pages 14788--14798, 2025.

\bibitem[Parkhi et~al.(2012)Parkhi, Vedaldi, Zisserman, and Jawahar]{parkhi2012cats}
Omkar~M Parkhi, Andrea Vedaldi, Andrew Zisserman, and CV Jawahar.
\newblock Cats and dogs.
\newblock In \emph{2012 IEEE conference on computer vision and pattern recognition}, pages 3498--3505. IEEE, 2012.

\bibitem[Ranftl et~al.(2021)Ranftl, Bochkovskiy, and Koltun]{ranftl2021vision}
Ren{\'e} Ranftl, Alexey Bochkovskiy, and Vladlen Koltun.
\newblock Vision transformers for dense prediction.
\newblock In \emph{Proceedings of the IEEE/CVF international conference on computer vision}, pages 12179--12188, 2021.

\bibitem[Shi et~al.(2025)Shi, Yan, Sun, Feng, Ren, Ma, Wang, Yin, de~Rijke, and Ren]{shi2025direct}
Zhengliang Shi, Lingyong Yan, Weiwei Sun, Yue Feng, Pengjie Ren, Xinyu Ma, Shuaiqiang Wang, Dawei Yin, Maarten de Rijke, and Zhaochun Ren.
\newblock Direct retrieval-augmented optimization: Synergizing knowledge selection and language models.
\newblock \emph{arXiv preprint arXiv:2505.03075}, 2025.

\bibitem[Shuai et~al.(2022)Shuai, Wu, and Liu]{shuai2022adaptive}
Hui Shuai, Lele Wu, and Qingshan Liu.
\newblock Adaptive multi-view and temporal fusing transformer for 3d human pose estimation.
\newblock \emph{IEEE Transactions on Pattern Analysis and Machine Intelligence}, 45\penalty0 (4):\penalty0 4122--4135, 2022.

\bibitem[Sim{\'e}oni et~al.(2025)Sim{\'e}oni, Vo, Seitzer, Baldassarre, Oquab, Jose, Khalidov, Szafraniec, Yi, Ramamonjisoa, Massa, Haziza, Wehrstedt, Wang, Darcet, Moutakanni, Sentana, Roberts, Vedaldi, Tolan, Brandt, Couprie, Mairal, J{\'e}gou, Labatut, and Bojanowski]{simeoni2025dinov3}
Oriane Sim{\'e}oni, Huy~V. Vo, Maximilian Seitzer, Federico Baldassarre, Maxime Oquab, Cijo Jose, Vasil Khalidov, Marc Szafraniec, Seungeun Yi, Micha{\"e}l Ramamonjisoa, Francisco Massa, Daniel Haziza, Luca Wehrstedt, Jianyuan Wang, Timoth{\'e}e Darcet, Th{\'e}o Moutakanni, Leonel Sentana, Claire Roberts, Andrea Vedaldi, Jamie Tolan, John Brandt, Camille Couprie, Julien Mairal, Herv{\'e} J{\'e}gou, Patrick Labatut, and Piotr Bojanowski.
\newblock {DINOv3}, 2025.

\bibitem[Singh et~al.(2021)Singh, Reddy, Hamilton, Dyer, and Yogatama]{singh2021end}
Devendra Singh, Siva Reddy, Will Hamilton, Chris Dyer, and Dani Yogatama.
\newblock End-to-end training of multi-document reader and retriever for open-domain question answering.
\newblock \emph{Advances in Neural Information Processing Systems}, 34:\penalty0 25968--25981, 2021.

\bibitem[Staal et~al.(2004)Staal, Abr{\`a}moff, Niemeijer, Viergever, and Van~Ginneken]{staal2004ridge}
Joes Staal, Michael~D Abr{\`a}moff, Meindert Niemeijer, Max~A Viergever, and Bram Van~Ginneken.
\newblock Ridge-based vessel segmentation in color images of the retina.
\newblock \emph{IEEE transactions on medical imaging}, 23\penalty0 (4):\penalty0 501--509, 2004.

\bibitem[Tschandl et~al.(2018)Tschandl, Rosendahl, and Kittler]{tschandl2018ham10000}
Philipp Tschandl, Cliff Rosendahl, and Harald Kittler.
\newblock The ham10000 dataset, a large collection of multi-source dermatoscopic images of common pigmented skin lesions.
\newblock \emph{Scientific data}, 5\penalty0 (1):\penalty0 1--9, 2018.

\bibitem[Wah et~al.(2011)Wah, Branson, Welinder, Perona, and Belongie]{wah2011caltech}
Catherine Wah, Steve Branson, Peter Welinder, Pietro Perona, and Serge Belongie.
\newblock The caltech-ucsd birds-200-2011 dataset.
\newblock 2011.

\bibitem[Xiao et~al.(2010)Xiao, Hays, Ehinger, Oliva, and Torralba]{xiao2010sun}
Jianxiong Xiao, James Hays, Krista~A Ehinger, Aude Oliva, and Antonio Torralba.
\newblock Sun database: Large-scale scene recognition from abbey to zoo.
\newblock In \emph{2010 IEEE computer society conference on computer vision and pattern recognition}, pages 3485--3492. IEEE, 2010.

\bibitem[Yu et~al.(2024)Yu, Tang, Xu, Cui, Ran, Yan, Liu, Wang, Han, Liu, et~al.]{yu2024visrag}
Shi Yu, Chaoyue Tang, Bokai Xu, Junbo Cui, Junhao Ran, Yukun Yan, Zhenghao Liu, Shuo Wang, Xu Han, Zhiyuan Liu, et~al.
\newblock Visrag: Vision-based retrieval-augmented generation on multi-modality documents.
\newblock \emph{arXiv preprint arXiv:2410.10594}, 2024.

\bibitem[Zhao et~al.(2025)Zhao, Chen, Chen, Liu, Chen, and Sun]{zhao2025retrieval}
Lin Zhao, Xiao Chen, Eric~Z Chen, Yikang Liu, Terrence Chen, and Shanhui Sun.
\newblock Retrieval-augmented few-shot medical image segmentation with foundation models.
\newblock \emph{IEEE Transactions on Neural Networks and Learning Systems}, 2025.

\bibitem[Zheng et~al.(2025)Zheng, Weng, Lyu, Jiang, Xue, Ren, Paudel, Sebe, Van~Gool, and Hu]{zheng2025retrieval}
Xu Zheng, Ziqiao Weng, Yuanhuiyi Lyu, Lutao Jiang, Haiwei Xue, Bin Ren, Danda Paudel, Nicu Sebe, Luc Van~Gool, and Xuming Hu.
\newblock Retrieval augmented generation and understanding in vision: A survey and new outlook.
\newblock \emph{arXiv preprint arXiv:2503.18016}, 2025.

\end{thebibliography}
}


\end{document}